\documentclass[journal]{IEEEtran}
\usepackage[utf8]{inputenc}

\ifCLASSINFOpdf
\else
   \usepackage[dvips]{graphicx}
\fi
\usepackage{url}

\usepackage{graphicx}
\usepackage{enumitem}
\usepackage{amsmath}
\usepackage{amsfonts}
\usepackage{amssymb}
\usepackage{booktabs}
\usepackage{caption}
\usepackage{subcaption}
\usepackage{siunitx} 
\usepackage{float}
\newcounter{subparagraph}[paragraph]
\renewcommand{\thesubparagraph}{\theparagraph.\arabic{subparagraph}}
\newcommand{\subparagraph}[1]{%
  \stepcounter{subparagraph}%
  \paragraph*{#1 (\thesubparagraph)}%
}

\begin{document} 

\title{4D Virtual Imaging Platform for Dynamic Joint Assessment via Uni-Plane X-ray and 2D–3D Registration }

\author{
  Hao Tang\textsuperscript{1,†}, 
  Rongxi Yi\textsuperscript{2,†}, 
  Lei Li\textsuperscript{1,†}, 
  Kaiyi Cao\textsuperscript{2}, 
  Jiapeng Zhao\textsuperscript{3}, 
  Yihan Xiao\textsuperscript{2}, 
  Minghai Shi\textsuperscript{2}, 
  Peng Yuan\textsuperscript{3}, 
  Yan Xi\textsuperscript{2}, 
  Hui Tang\textsuperscript{3}, 
  Wei Li\textsuperscript{2,*}, 
  Zhan Wu\textsuperscript{3,*}, 
  Yixin Zhou\textsuperscript{1,*}

    \textsuperscript{1}\,Department of Adult Joint Reconstructive Surgery, Beijing Jishuitan Hospital, Capital Medical University, Beijing, China \\

    \textsuperscript{2}\,First Imaging Medical Equipment, Shanghai, China \\
    \textsuperscript{3}\,Key Laboratory of New Generation Artificial Intelligence Technology and Its Interdisciplinary Applications,\\
    Ministry of Education, Southeast University, Nanjing, China \\

    \textsuperscript{†}\,These authors contributed equally to this work (co-first authors). \\
    \textsuperscript{*}\,Corresponding authors: weili@first-imaging.com, 101300254@seu.edu.cn, orthoyixin@gmail.com

}



    



\maketitle
\begin{abstract}

Conventional computed tomography (CT) lacks the capability to capture dynamic, weight-bearing joint motion. Functional assessment, especially after surgical intervention, requires four-dimensional (4D) imaging, yet current methods are constrained by excessive radiation or incomplete spatial information from 2D techniques.  We propose an integrated 4D joint analysis platform that unites: (1) a dual–robotic-arm CBCT system with a programmable, gantry-free trajectory optimized for upright scanning; (2) a hybrid imaging pipeline that fuses static 3D CBCT with dynamic 2D X-rays using deep learning–based preprocessing, 3D–2D projection, and iterative optimization; and (3) a clinically validated framework for quantitative kinematic assessment. In simulation studies, the method achieved sub-voxel accuracy (0.235~mm) with a 99.18\% success rate, outperforming conventional and state-of-the-art registration approaches. Clinical evaluation further demonstrated accurate quantification of tibial plateau motion and medial–lateral variance in post–total knee arthroplasty (TKA) patients. This 4D CBCT platform enables fast, accurate, and low-dose dynamic joint imaging, offering new opportunities for biomechanical research, precision diagnostics, and personalized orthopedic care.  

\end{abstract}


\IEEEpeerreviewmaketitle

\section{Introduction}

Since the introduction of 2D radiography in the 19th century \cite{2021_Godoy-Santos, VanTiggelen2001_HistoryXray, wu2025cone}, orthopedic imaging technologies have advanced considerably. Conventional computed tomography (CT) and magnetic resonance imaging (MRI) provide accurate three-dimensional (3D) anatomical assessment but require patients to be scanned in a horizontal, non–weight-bearing position, limiting their ability to capture the true anatomical relationships of bones and joints under physiological load. In contrast, 2D radiography can be performed in weight-bearing positions but is limited by projection distortion and structural overlap \cite{wu2025praise,yuan2025multi}.

Weight-bearing cone-beam CT (CBCT) has emerged as a promising alternative, enabling 3D imaging of the foot, ankle, and knee in a load-bearing, muscle-activated state \cite{diagnostics14151641,spine_stability_cbct}. This modality facilitates evaluation of joint alignment and load distribution under physiological conditions, offering unique value for diagnosis, treatment planning, and postoperative assessment. However, CBCT remains restricted to \textit{static structural imaging} and cannot capture temporal information. Functional evaluation—particularly for assessing the outcomes of surgical interventions—requires temporal 3D (4D) imaging to characterize both anatomy and motion \cite{eck2002biomechanical,wu2023deep}. At present, temporal data are primarily derived from conventional 2D imaging, which lacks full spatial context, whereas dynamic 3D CT is clinically impractical due to excessive radiation exposure.  

Recent advances in 2D–3D registration have attempted to bridge static volumetric data with dynamic imaging. Classical iterative algorithms (e.g., Powell \cite{10.1093/comjnl/7.2.155}, Nelder–Mead \cite{Nelder1965ASM}) provide robustness but are computationally intensive, while deep learning–based methods (e.g., PEHL \cite{PEHL}, ProST \cite{ProST}) achieve greater efficiency and extended capture ranges but often require large, task-specific datasets and are sensitive to domain shifts. These trade-offs highlight the need for approaches that combine the accuracy and robustness of iterative strategies with the efficiency of learning-based techniques. 

To address these limitations, we propose an integrated 4D joint anatomical analysis platform that combines a novel weight-bearing imaging system, optimized acquisition protocols, advanced registration algorithms, and clinically validated evaluation tools. 
The imaging system is a dual–robotic-arm CBCT, designed with two primary objectives: (1) to enable scanning in natural, weight-bearing positions, and (2) to adopt a gantry-free configuration that accommodates greater patient motion and facilitates integration with other biomedical sensors\cite{lin2024full}. Compared with other CBCT architectures—such as C-arm and O-arm systems—the dual–robotic-arm configuration offers superior flexibility in trajectory programming, providing optimal coverage for motion capture.  Building upon this system, we developed a time-efficient and robust 4D imaging pipeline. The acquisition protocol begins with a static 3D CBCT scan, followed by dynamic 2D X-ray sequences capturing joint motion. Performing the static scan first minimizes relative displacement between the 3D and 2D datasets, simplifying registration. A hybrid 2D–3D registration strategy is then applied, integrating deep learning–based preprocessing, 3D–2D projection, and conventional iterative optimization \cite{imu_motion_cbct}. This approach achieves high accuracy and robustness with reduced data preparation requirements, while operating more efficiently than purely iterative methods.  

To ensure that kinematic measurements translate into clinically actionable findings, we collaborated with orthopedic specialists to design a systematic evaluation framework \cite{lintz2021weight,buzzatti2023dynamic}. Key metrics include tibial plateau (TP) and condyle–TP distances, medial–lateral distance differences, medial/lateral projection point distributions, and medial/lateral projection linkage plots. These quantitative measures, when interpreted alongside clinical records, provide insight into bone wear, ligament balance, and deformities such as genu varum or valgum, thereby supporting diagnosis and enabling personalized treatment planning.  

In summary, this study introduces an integrated 4D CBCT joint anatomical analysis platform that addresses the limitations of existing imaging modalities in capturing dynamic, weight-bearing motion at low radiation dose. By combining a dual–robotic-arm CBCT system optimized for physiological scanning, a hybrid 4D imaging pipeline, and a clinically validated evaluation framework, we provide orthopedic specialists with a precise and robust tool for motion analysis—bridging the gap between static anatomical visualization and dynamic functional assessment.





\section{\textbf{Weight-bearing dual-robotic-arm CBCT}}
\begin{figure}
    \centering
    \includegraphics[width=1\linewidth]{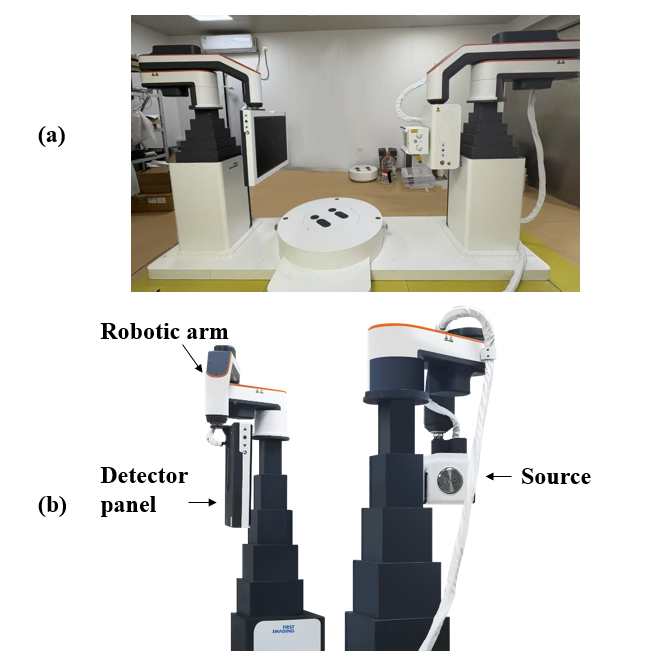}
    \caption{Weight-bearing dual-robotic-arm CBCT system}
    \label{fig:weight-bearing CBCT system}
\end{figure}

We developed a gantry-free, dual–robotic-arm CBCT system (First-Imaging Jishuitan system, Figure~\ref{fig:weight-bearing CBCT system}(a)), in which two ultra-lightweight robotic arms (16\,kg each) independently control the X-ray source and flat-panel detector (Figure~\ref{fig:weight-bearing CBCT system}(b)). Mounted on a stable base, the arms eliminate the need for ceiling reinforcement while providing greater flexibility than conventional CBCT systems \cite{lin2024full}. With integrated vertical lift and rotary bearings, the system achieves a maximum z-axis travel of 180\,cm, sufficient for comprehensive musculoskeletal evaluation in upright positions. The 61-cm-wide detector provides a broad field of view, well-suited for orthopedic applications. Multi–degree-of-freedom arm movement allows trajectory adjustment to accommodate patient anatomy, avoid obstructions such as orthotic devices, and optimize projection angles for stable single-plane 2D–3D registration \cite{KUWASAWA2023408}. The load-bearing imaging allows for a more accurate assessment of the stress distribution in the knee, foot, and ankle, and many musculoskeletal disorders, particularly those of the spine and lower extremities, become more evident under such conditions \cite{spine_stability_cbct,KUWASAWA2023408,Geng2023}.

Subjects are scanned in natural standing or semi-flexed postures on a weight-bearing platform. The dual–robotic-arm CBCT performs a 360$^{\circ}$ scan, and projections are reconstructed into high-resolution volumes using an enhanced Feldkamp–Davis–Kress (FDK) algorithm with reverse-spiral trajectories \cite{yu2011fdk}. Projection-weight normalization compensates for uneven sampling, yielding superior image quality compared with standard FDK \cite{lin2024full}. These 3D volumes capture load-dependent features such as joint surface contact, cartilage thickness, and osteophyte distribution that are difficult to assess in non-weight-bearing states. In parallel, a high-frequency lateral X-ray system records fluoroscopic sequences while subjects maintain upright posture and perform controlled motions (e.g., knee flexion, leg lifting, stair simulation), providing real-time trajectories of joint surfaces and joint-space changes for 2D–3D registration and virtual 4D reconstruction \cite{range_motion_cbct}. To validate the pipeline, synthetic X-rays are generated from reconstructed CBCT volumes using a system-specific ray-driven projection algorithm that incorporates the scanner’s trajectory and attenuation profile, producing realistic projection data with known ground truth for both clinical evaluation and algorithm benchmarking.


\section{\textbf{4D virtual Imaging}}

Based on the dual–robotic-arm CBCT system \cite{lin2024full}, we developed a 4D imaging pipeline comprising data acquisition, pose initialization, anatomical segmentation, and image registration. The pipeline fuses dynamic 2D X-ray sequences with static 3D CBCT volumes to generate high-precision, time-resolved 4D reconstructions for quantitative kinematic assessment. The most essential part of 4D virtual imaging is 2D-3D registration which (Figure \ref{fig:registration_workflow}) consists of three stages: multiobject segmentation, simulated 3D-2D projection and iterative registration.


\begin{figure}[ht]
\centering
\includegraphics[width=1\linewidth]{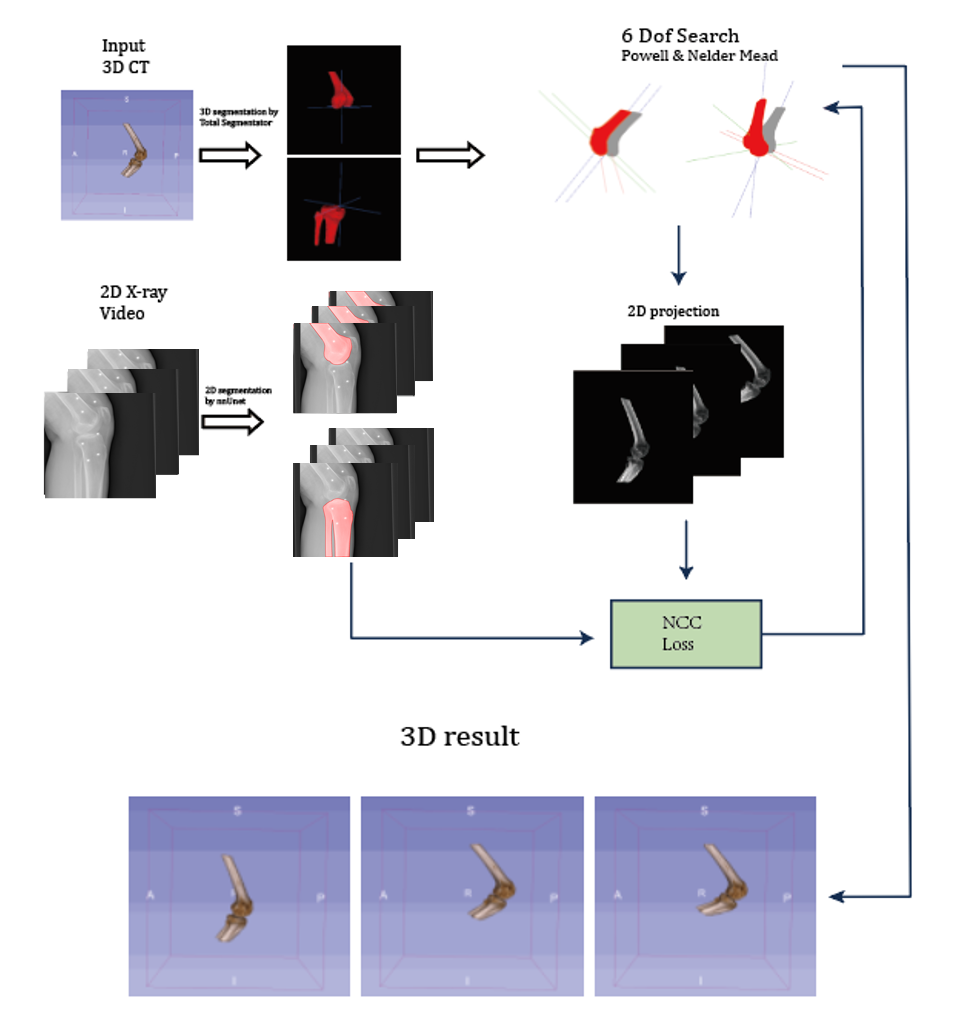}
\caption{Framework of 4D Virtual Imaging }
\label{fig:enter-label}
\end{figure}


\subsection{\textbf{Multi-object Segmentation}} 
Accurate delineation of skeletal structures is critical for both registration and orthopedic analysis. We focused on three key components of the knee joint—the femur, patella, and tibia–fibula complex—as they capture the primary biomechanics and are standard targets in both radiographic and volumetric imaging.

For 2D X-ray images, segmentation was performed using an adapted nnU-Net framework \cite{NNUnet,isensee2021nnunet}. Unlike conventional frame-independent approaches, our model incorporates temporal consistency: the first frame is manually refined, and subsequent frames are segmented with reference to the preceding results. By combining sequential information with 2D registration cues, this strategy achieves higher accuracy in dynamic motion sequences while retaining nnU-Net’s strengths in adaptability and automated configuration across diverse joint motions. For 3D CBCT volumes segmentation, TotalSegmentator \cite{wasserthal2023totalsegmentator} was selected as a versatile CT segmentation solution for various bone structures, including the femur, patella, and tibia-fibula complex.


\subsection{\textbf{Deep simulated 3D-2D projection}} 

For 2D-3D registration, CBCT volumes are projected into 2D space using the DeepDRR framework \cite{unberath2018deepdrrcatalystmachine}, which generates realistic Digitally Reconstructed Radiographs (DRRs) from diagnostic CT. DeepDRR integrates four modules: a CNN for material decomposition and segmentation, a spectrum-aware ray-tracing projector to build attenuation maps, a CNN-based scatter estimator, and noise simulation for quantum and electronic readout effects. Compared with direct CBCT projections, DeepDRR yields fluoroscopy images that more closely match clinical data, minimizing mismatch during registration. More importantly, it produces results within seconds regardless of photon count, improving registration accuracy without sacrificing computational efficiency \cite{maier2020review}.  


\subsection{\textbf{Iterative 2D-3D Registration}} 

Subsequent to segmentation of 2D X-ray bone structures and 3D CBCT anatomy, a 2D–3D registration framework is applied to align the segmented 2D skeletal regions with the forward projections of their corresponding 3D CBCT models \cite{maier2020review,xu2021learningbased}. Each target bone is modeled as a rigid body undergoing six degrees of freedom (6 DoF) transformations:
\begin{equation}
T = { t_i = (t_{xi}, t_{yi}, t_{zi}, r_{\alpha i}, r_{\beta i}, r_{\gamma i}) \ | \ i = 1,2,3 }
\end{equation}
where translational components \(t_{xi}\), \(t_{yi}\), \(t_{zi}\) and rotational angles \(r_{\alpha i}\), \(r_{\beta i}\), \(r_{\gamma i})\) collectively define the rigid transformation.

This registration process maps 3D bone models into the pose space of 2D X-ray images, enabling rigid alignment of anatomical structures across modalities while preserving spatial consistency. Using Tong’s iterative projection–matching method \cite{lin2024full}, segmented 2D contours are aligned with synthetic 3D projections to generate a sequence of pose-corrected 3D volumes. These frames collectively form a dynamic (4D) CBCT series, allowing non-invasive reconstruction of joint kinematics across continuous anatomical poses. The combination of robust segmentation and modality-specific registration ensures that the resulting 4D CBCT accurately reflects in-vivo skeletal orientations, providing a solid foundation for motion analysis, biomechanical modeling, and personalized treatment planning.

This stepwise design balances computational efficiency and registration accuracy, offering a robust solution for multi-modal alignment of complex joint anatomy. Detailed implementation is described in the following:

\begin{figure}
    \centering
    \includegraphics[width=1\linewidth]{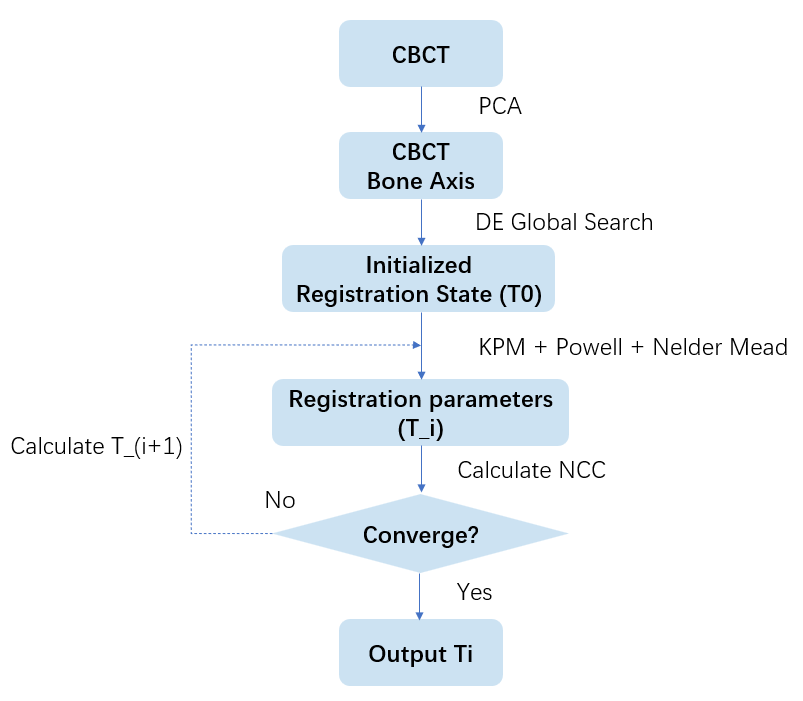}
    \caption{Our proposed registration workflow (BoneAxis-Reg)}
    \label{fig:registration_workflow}
\end{figure}

\subsubsection{\textbf{BoneAxis-Reg Framework}}  
The proposed three-step 2D–3D registration framework (BoneAxis-Reg) achieves spatial alignment by rotating CBCT volumes along bone axes under a hierarchical optimization strategy (Figure \ref{fig:registration_workflow}):

\begin{enumerate}[label=\alph*)]
    \item \textbf{Global initialization:} A differential evolution (DE) algorithm searches along each bone’s principal axis (derived via PCA), exploring a wide parameter space to identify accurate initialization and reduce the burden of subsequent refinement.  
    \item \textbf{Kinematic constraints:} A kinematic priority module (KPM) embeds physiological motion constraints into the optimization, narrowing the feasible solution space and accelerating convergence.  
    \item \textbf{Local refinement:} A hybrid optimizer combining the Powell method and Nelder–Mead simplex performs coarse-to-fine optimization. Global registration captures large-scale transformations, while local refinement resolves subtle discrepancies. Normalized cross-correlation (NCC) is used as the similarity metric.  
\end{enumerate}


\subsubsection{\textbf{Two-step Initialization of Registration Parameters}}  
1. The semantically segmented 3D bone volumes are transformed to a set of 3D  point clouds. Bone mechanical axes are then determined by applying principal component analysis (PCA) to point cloud data. 
2. With the femoral and tibio-fibular mechanical axes established in step 1, the differential evolution (DE) algorithm conducts a global exploration of the transformation space. This process iteratively mutates and recombines candidate solutions, leveraging the PCA-derived axes as the initial search seed to efficiently navigate the parameter landscape and identify optimal initial postures. Within this step, the initial position can be obtained in the form of:
\begin{equation}
T_0 = {(t_{x0}, t_{y0}, t_{z0},r_{\alpha 0}, r_{\beta 0}, r_{\gamma 0}) }
\end{equation}
where $T_0$ stands for the initial 6 DoF parameters of registration (translation in x,y,z and rotation in $\alpha$, $\beta$, $\gamma$) to be used as the starting point of the following registrations.

\subsubsection{\textbf{Kinematic Priority Module}}  
To further improve the efficiency of registration optimization, we propose a kinematic priority module (KPM) that incorporates the priority of our clinical knee joint movement protocol and employs kinematic parameters as spatiotemporal constraints to refine the registration optimization process.
For simplifying the modeling of knee joint movement, we only capture the moments of knee extension and flexion, which can be described by the following mathematical formula::
Thus, the registration optimization problem can be simplified into solving the equation: 
\begin{equation}
T_n = T_{n-1} + KPM * S
\end{equation}
where $T_n$ stands for the 6 DoF parameters of the $n_{th}$ iteration, KPM stands for KPM optimization direction and $S$ stands for the step size of KPM.

\subsubsection{\textbf{Normalized Cross-Correlation Optimization}}  
The objective is to maximize the similarity between the synthetic projection and the 2D X-ray image, thereby ensuring accurate alignment of bone contours in both images. This similarity can be quantified using the normalized cross-correlation (NCC) metric. The NCC between two images $I_1$ and $I_2$ over a region $\Omega$ is given by:


\begin{equation}
\text{NCC}(I_1, I_2) = \frac{\sum (I_1 - \bar{I}_1)(I_2 - \bar{I}_2)}{
\sqrt{\sum (I_1 - \bar{I}_1)^2} \sqrt{\sum (I_2 - \bar{I}_2)^2}}
\end{equation}
where $\bar{I}_1$ and $\bar{I}_2$ represent the mean intensities of $I_1$ and $I_2$, respectively. The NCC value ranges from -1 (anti-correlation) to 1 (perfect correlation).

To achieve the fastest convergence speed and highest accuracy, a hybrid optimization strategy is proposed that synergistically integrates the Powell method and the Nelder-Mead method. This strategy capitalizes on their complementary strengths to enhance both global exploration and local exploitation capabilities.


\section{Experiments and Datasets}

To evaluate the effectiveness of the proposed 4D CBCT framework, three datasets were employed: a real lamb joint dataset, a simulated human joint dataset, and human clinical data.  

The \textbf{lamb joint dataset} enabled quantitative accuracy assessment. Because bone-fixed markers could be rigidly attached without invasive procedures, reliable ground-truth kinematics were obtained.  

For humans, however, ground-truth motion is difficult to measure. Skin-attached markers are subject to soft tissue artifacts, and invasive bone markers are clinically impractical. To address this, a \textbf{simulated human dataset} was generated with known kinematic conditions for quantitative validation.  

Finally, the \textbf{clinical dataset} was used to demonstrate in vivo feasibility. The reconstructed kinematics were qualitatively cross-validated against medical records and clinical observations.  

\subsubsection{\textbf{Lamb Knee Joint Dataset}}  
An adult lamb hindlimb (2.3 kg) was prepared, preserving the femur–tibia–fibula complex and associated structures:  

- \textbf{Skeletal system}: cancellous and cortical bone of the distal femur and proximal tibia–fibula, articular cartilage layers (1.2–1.5 mm thick), and the patellofemoral complex.  

- \textbf{Soft tissue structures}: quadriceps and hamstring tendon attachments, medial/lateral collateral ligaments, anterior/posterior cruciate ligaments, and the joint capsule, maintaining muscle–tendon continuity (muscle weight ratio 35 ± 5 \%).  

Non-essential tissues (e.g., subcutaneous fat) were removed to minimize interference while retaining physiological load-transfer properties.  

Three standardized poses were defined:  
a) \textbf{Neutral extension}: 0° flexion with 5° femoral internal rotation.  
b) \textbf{Active flexion}: 60° ± 5° flexion, simulating the weight-bearing phase of gait.  
c) \textbf{Valgus stress}: 10° abduction, simulating medial collateral insufficiency.  

\begin{figure}
    \centering
    \includegraphics[width=0.6\linewidth]{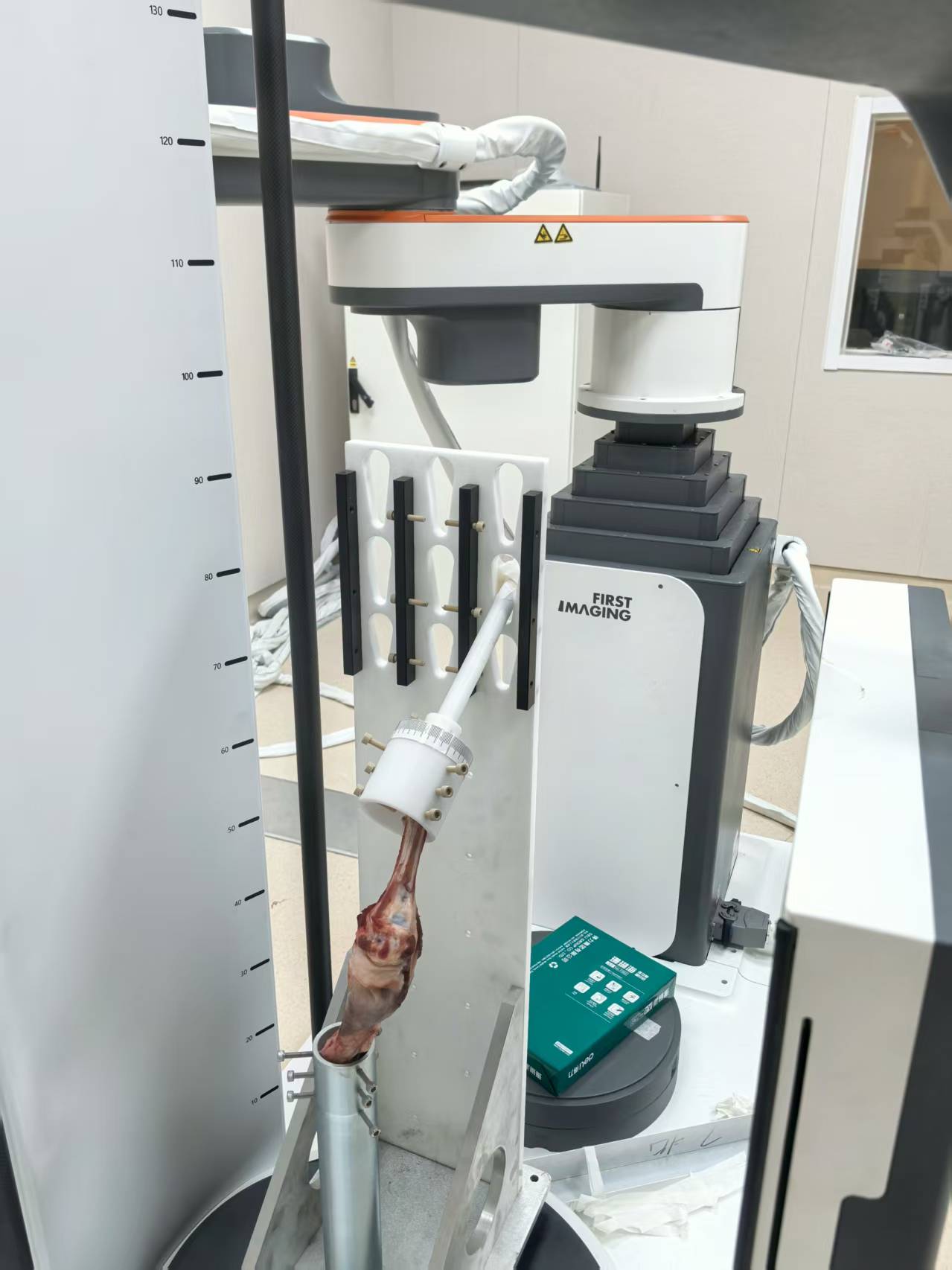}
    \caption{Lamb knee joint clinical trial scenario}
    \label{fig:Lamb leg experiment}
\end{figure}

Static 3D CBCT volumes were acquired using a high-precision scanner (First-Imaging Jishuitan System), with corresponding 2D fluoroscopy images (pixel spacing 2.1 mm, resolution 1024 × 1024), as shown in Figure \ref{fig:Lamb leg experiment}.  

A custom fixation device allowed the joint to be locked in discrete “frozen frames” of motion. For each frame, a 3D CBCT scan was acquired with bone-fixed external markers. The system was calibrated to localize markers with <0.1 mm error, verified by optical navigation. These marker-derived measurements provided ground truth against which 2D–3D registration from the 4D CBCT pipeline was quantitatively validated.  

\subsubsection{\textbf{Simulated Human Joint Dataset}}  
Ten frames were synthesized from a static human CBCT volume by mimicking natural knee flexion–extension. Dynamic 2D X-ray projections were generated from these pose-specific bone models using Tong’s method \cite{lin2024full}, incorporating realistic noise and geometric distortions to emulate clinical conditions.  

\subsubsection{\textbf{Human Clinical Data}}  
One post-TKA patient with indications for osteotomy or arthroplasty and capable of weight-bearing during imaging was included.  

Static 3D CBCT scans (1024 × 1024, 110 kVp, 51 mA, 15 ms) were acquired with a weight-bearing CBCT system (First-Imaging Jishuitan System). A 180° reverse spiral trajectory (1.5 mm step) covered the femur–tibia–fibula complex, with robotic vibration corrected via steel-ball calibration. High-speed single-plane fluoroscopy images were acquired during active knee flexion at 0°, 30°, 60°, and 90° using the same device (0.3 mm pixel size, 1024 × 1024 resolution).

\section{\textbf{Evaluation and results}}

\subsection{\textbf{Evaluation Metrics for Registration}}  
To quantitatively evaluate the proposed method, the following metrics were employed:  

\textbf{6-Degree-of-Freedom (6-DoF) Registration Error:}  
The 6-DoF error is computed as the absolute difference between the estimated transformation parameters and the ground-truth transformation ($T_{gt}$). For each degree of freedom (three translations and three rotations), absolute errors are calculated across all test cases and then averaged. This metric provides a comprehensive measure of alignment accuracy across spatial dimensions.

\textbf{Target Registration Error (TRE):}  
TRE assesses spatial consistency by measuring the average Euclidean distance between corresponding points on the two transformed volumes. It is formally defined as:  
 
\begin{equation}
    TRE = \sum_{s=1}^{N_{BonePixel}} \frac{\| S_i \cdot \mathcal{T}(T_{\text{est}}) - S_i \cdot \mathcal{T}(T_{gt}) \|_2}{N_{BonePixel}} 
    \label{eq:placeholder_label}
\end{equation}
where $T_{gt}$ denotes the ground-truth transformation, $T_{est}$ the estimated transformation, $T$ the spatial transformation operator, and $S_i$ the set of points within the bone mask. This metric quantifies the geometric discrepancy between the predicted and the real transformations in 3D space, reflecting the practical accuracy of point registration on anatomical structures.  

\textbf{Registration Success Rate (RSR):}  
Based on clinical evidence, biomechanical stability, and surgical feasibility in TKA preoperative planning, we adopted 1.5 mm as the TRE threshold for defining registration success. Prior studies \cite{Levy2017, Murat2017} identify this value as the upper bound of acceptable error, with most resections using patient-specific guides remaining within it. Errors exceeding 1.5 mm increase the risk of joint malalignment, soft tissue imbalance, and accelerated implant wear, thereby compromising long-term stability.  

Mechanically, this threshold ensures balanced knee load distribution and reduces stress concentrations that may lead to component failure. Anatomically, it preserves joint symmetry and minimizes postoperative instability. Cadaveric and clinical data further show that deviations within 1.5 mm rarely require intraoperative correction, aligning well with modern navigation and 3D-printed guide techniques. Thus, 1.5 mm represents a practical balance between anatomical precision and surgical reality, optimizing both short-term outcomes and long-term durability.  

For the lamb dataset, a relaxed 3 mm threshold was adopted, accounting for smoother bone surfaces and fewer distinct anatomical landmarks compared with the human knee.   

\subsection{\textbf{Registration Accuracy}}  

To evaluate registration performance, four state-of-the-art registration frameworks were selected as baselines: Nelder-Mead–NCC, Powell + Nelder-Mead–NCC, PEHL, and ProST.  

\begin{itemize}  
\item \textbf{Nelder-Mead:} A derivative-free optimizer that iteratively adjusts a simplex through geometric operations (reflection, expansion, contraction) to minimize similarity metrics (e.g., mutual information). It requires no gradient computation, is easy to implement, and remains robust with noisy or low-dimensional data, making it suitable for cross-modality registration \cite{Nelder1965ASM}.  

\item \textbf{Powell:} A gradient-free method that minimizes similarity metrics (e.g., normalized mutual information) via conjugate direction search \cite{10.1093/comjnl/7.2.155}. It offers fast convergence and high precision for rigid/non-rigid alignment (e.g., CT–MRI), but is sensitive to initialization and prone to local minima, often necessitating combination with global optimizers such as Nelder–Mead.  

\item \textbf{PEHL:} Miao et al. proposed a CNN-based hierarchical regression approach (PEHL)  that decomposes registration tasks into subtasks and directly estimates transformation parameters from residuals between digitally reconstructed radiographs (DRRs) and X-rays, achieving real-time performance with accuracy comparable to intensity-based methods \cite{PEHL}.

\item \textbf{ProST:} A neural network–based 2D–3D registration framework that learns gradient similarity between source and target images. During registration, iterative optimization is guided by the pre-trained network, enabling robust pose estimation (translation/rotation) under complex imaging conditions \cite{markelj2008robust}.  
\end{itemize}  

\noindent\textbf{Simulated human knee data.}  
As shown in Table~\ref{tab:registration_error_human_simulated_1.5mm}, the proposed method consistently achieves lower errors across all six degrees of freedom (DoF). Its target registration error (TRE) is 0.948 mm, outperforming Nelder–Mead (1.559 mm) and Powell + Nelder–Mead (6.778 mm). With a 1.5 mm RSR threshold, our method achieves 87.57\% success, compared with 69.03\% for Nelder–Mead and 43.54\% for Powell + Nelder–Mead, confirming its superior accuracy on simulated human data.  

\noindent\textbf{Real lamb joint data.}  
For real joint experiments, our method again surpasses baseline frameworks across all metrics. It achieves the lowest 6 DoF registration error, indicating more accurate alignment of anatomical structures in all motion axes. TRE (1.338 mm)  and RSR (90.84\%) results further demonstrate greater robustness and reliability compared with existing approaches.

\begin{table*}[]
    \centering
    \caption{6-DoF Registration Error, TRE and RSR for Simulated Human Knee, RSR Threshold=1.5mm}
    \begin{tabular}{lcccccccc}
        \hline
        \textbf{Method} & \textbf{$t_x$(mm)} & \textbf{$t_y$(mm)} & \textbf{$t_z$(mm)} & \textbf{$r_{\alpha}(^\circ)$} & \textbf{$r_{\beta}(^\circ)$} & \textbf{$r_{\gamma}(^\circ)$} & \textbf{TRE(mm)} & \textbf{RSR(\%)} \\
        \hline
        BoneAxis-Reg& 0.124& 0.116& 1.685& 0.369& 0.215& 0.158& 0.948& 87.57\\
 BoneAxis-Reg + KPM& 0.014& 0.010& 0.012& 0.039& 0.031& 0.013& 0.235&99.18\\
        Nelder Mead& 0.282& 0.442& 2.993& 2.693& 3.092& 0.800& 1.559& 69.03\\
        Powell + Nelder Mead& 2.127& 2.0396& 13.475& 1.970& 2.675& 1.279& 6.778& 43.54\\
        PEHL& 2.440& 1.460& 1.455& 1.015& 1.370& 1.215& 1.316& 83.32\\
        ProST & 7.930& 15.015& 1.970& 0.680& 1.200& 4.760& 6.321& 23.62\\
        \hline
    \end{tabular}

    \label{tab:registration_error_human_simulated_1.5mm}
\end{table*}




\begin{table*}[]
    \centering    
    \caption{Average Translation ($mTrans$) and Rotation ($mRot$) Errors for Simulated Human Knee.}
    \begin{tabular}{|l|c|c|}
        \hline
        \textbf{Method}

        & \multicolumn{2}{c|}{\textbf{Simulated Human Knee}} 
         \\
        \cline{2-3} 
        & $mTrans$(mm) & $mRot(^\circ)$ \\
        \hline
        BoneAxis-Reg
& 0.642& 0.247\\
        BoneAxis-Reg + KPM
& 0.012& 0.027\\
 Nelder Mead
& 1.239& 2.195\\
        Powell + Nelder Mead
& 5.880& 1.975\\
        PEHL& 1.785& 1.200\\
        ProST & 8.305& 2.213\\
        \hline
    \end{tabular}

    \label{tab:comparison_total_human_simulated}
\end{table*}

\begin{table*}[]
    \centering
    \caption{6-DoF Registration Error, TRE and RSR for Real Lamb knee joint, RSR Threshold=3mm}
    \begin{tabular}{lcccccccc}
        \hline
        \textbf{Method} & \textbf{$t_x$(mm)} & \textbf{$t_y$(mm)} & \textbf{$t_z$(mm)} & \textbf{$r_{\alpha}(^\circ)$} & \textbf{$r_{\beta}(^\circ)$} & \textbf{$r_{\gamma}(^\circ)$} & \textbf{TRE(mm)} & \textbf{RSR(\%)} \\
        \hline
        BoneAxis-Reg& 0.420& 0.508& 5.745& 1.689& 1.485& 0.101& 1.338& 90.84\\
        Nelder Mead& 7.676& 11.857& 16.503& 12.494& 14.776& 15.010& 11.850& 21.13\\
        Powell + Nelder Mead& 9.253& 9.517& 18.940& 3.757& 11.914& 12.089& 10.160& 17.51\\
        PEHL& 6.360& 11.890& 2.555& 1.705& 1.075& 4.205& 2.838& 54.83\\
        ProST & 1.465& 1.220& 6.120& 4.335& 4.010& 0.985& 2.730& 56.22\\
        \hline
    \end{tabular}

    \label{tab:registration_error_real_lamb_3mm}
\end{table*}

\subsection{\textbf{Impact of the Kinematic Priority Module}}  

Table~\ref{tab:registration_error_human_simulated_1.5mm} compares results from the proposed method with and without the kinematic priority module (KPM) on simulated human knee data. Per-frame translation and rotation errors show that KPM markedly reduces both translational and rotational discrepancies. Specifically, the TRE decreases from 0.948 mm to 0.235 mm, while the RSR improves from 87.57\% to 99.18\%. These results demonstrate that incorporating kinematic constraints into the optimization process yields substantially more accurate registration.  

Additional evidence is provided in Table~\ref{tab:comparison_total_human_simulated}, which reports average translation and rotation errors. With KPM, the mean translation error (mTrans) decreases from 0.642 mm to 0.012 mm, and the mean rotation error (mRot) from 0.247° to 0.027°. This confirms that KPM not only refines alignment accuracy but also enhances robustness, making the framework well-suited for high-precision applications.  

\subsection{\textbf{Clinical Test on Real Human Knee Data}}  

For clinical validation, we tested the framework on a TKA patient with documented knee instability. During step-ascent motion under 2D dynamic X-ray imaging, multi-dimensional kinematic parameters—including medial–lateral difference (MLD), distance difference variance (DDV), and trajectory patterns—were extracted. These metrics provide clinically relevant insights into joint stability and abnormal motion, supporting the framework’s utility in real-world evaluation of post-operative outcomes.

\begin{figure}[]
    \centering
    \includegraphics[width=1\linewidth]{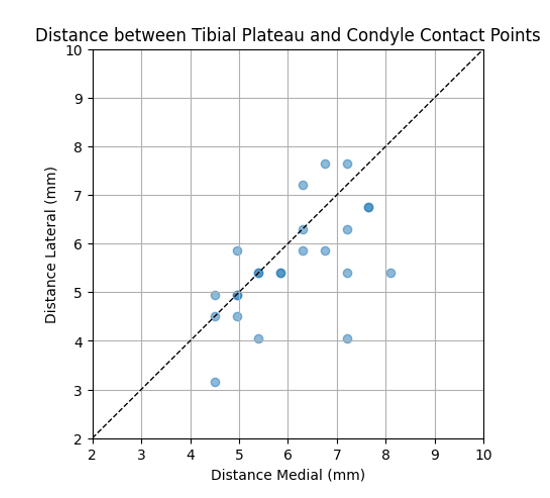}
    \caption{Tibial plateau-condyle contact points}
    \label{fig:TP Condyle contact points sub1}
\end{figure}

\begin{figure}
    \centering
    \includegraphics[width=1\linewidth]{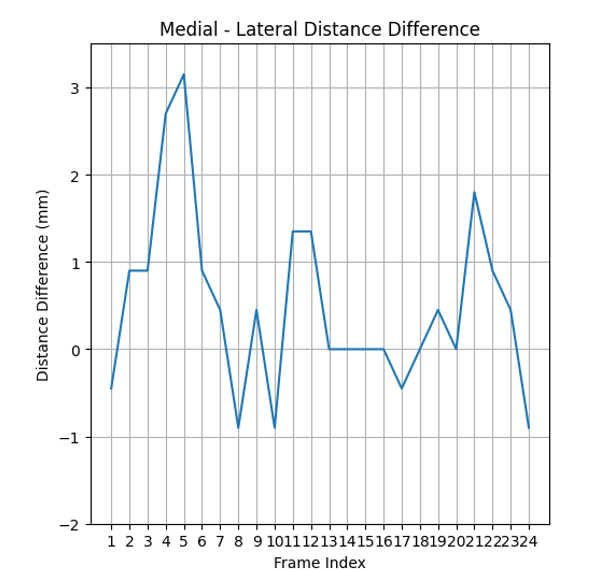}
    \caption{Media-lateral distance difference across knee motion frames}
    \label{fig:media-lateral difference sub1}
\end{figure}

\begin{figure}
    \centering
    \includegraphics[width=1\linewidth]{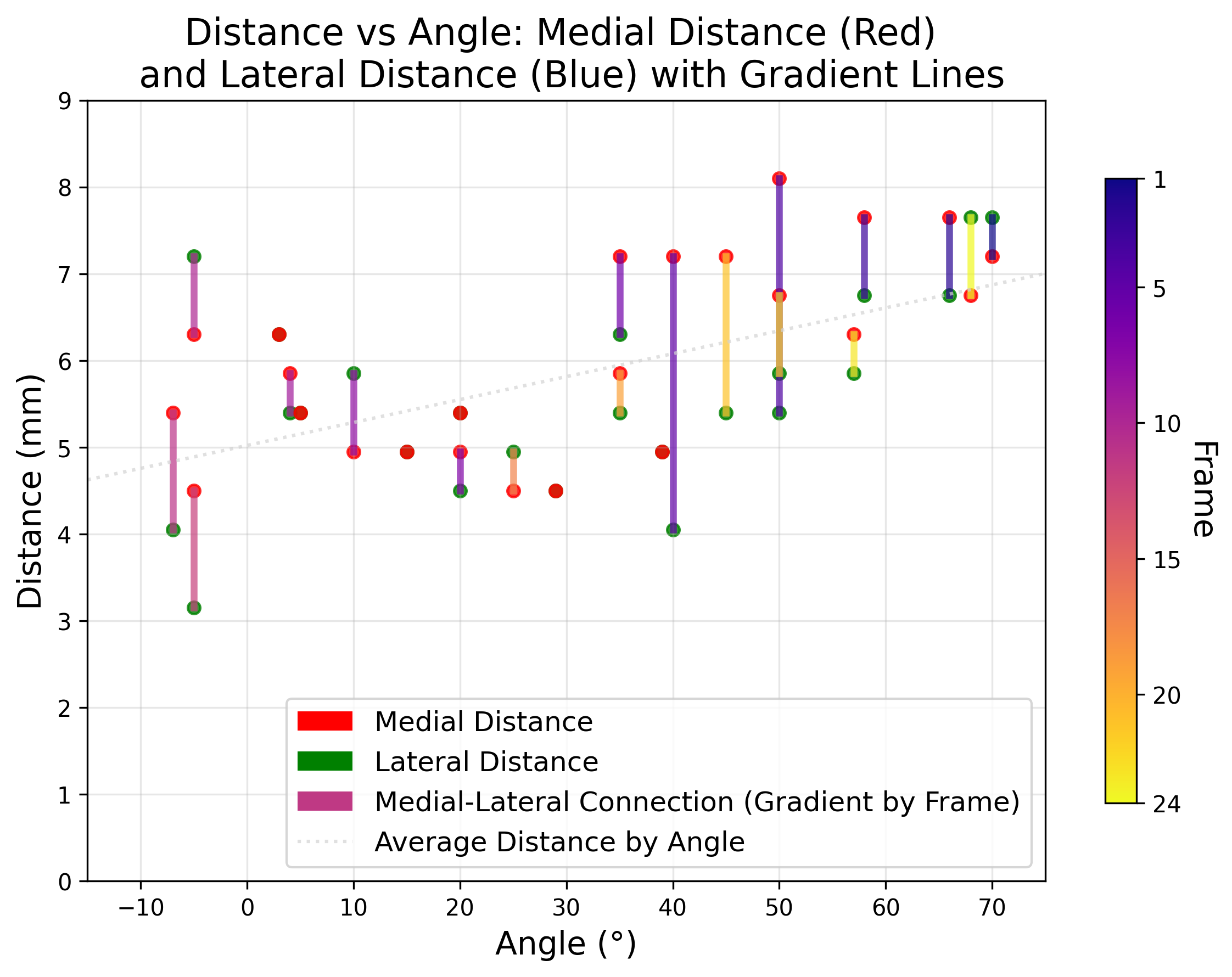}
    \caption{Knee extension angle VS medial/lateral distance}
    \label{fig:Extension angle VS medial/lateral distance sub1}
\end{figure}

\begin{figure}
    \centering
    \includegraphics[width=1\linewidth]{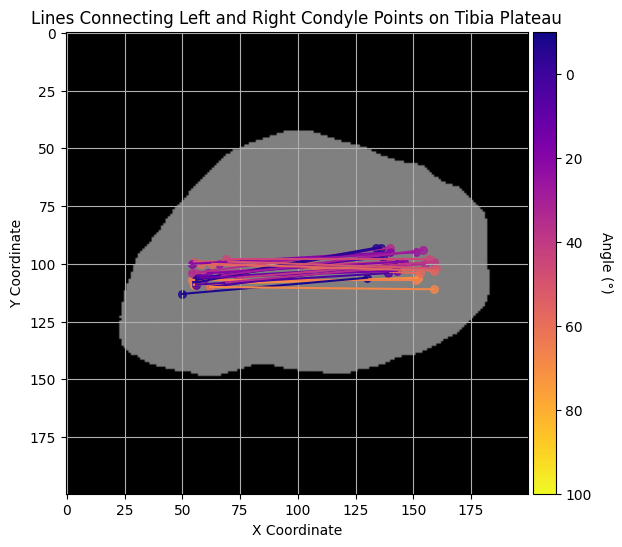}
    \caption{Trajectory analyses of knee joint motions}
    \label{fig:trajectory_analysis}
\end{figure}

\begin{figure}
    \centering
    \includegraphics[width=0.75\linewidth]{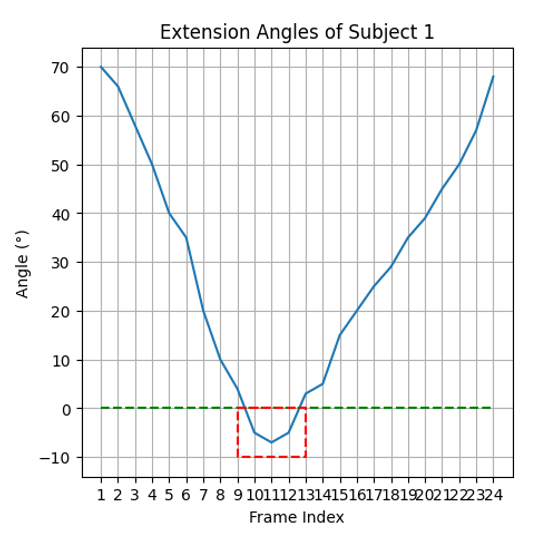}
    \caption{Knee motion frames and corresponding joint angles of the clinical subject}
    \label{fig:hyperextension_frames}
\end{figure}
 



\begin{figure}
    \centering
    \begin{subfigure}{1\linewidth}
        \centering
        \includegraphics[width=\textwidth]{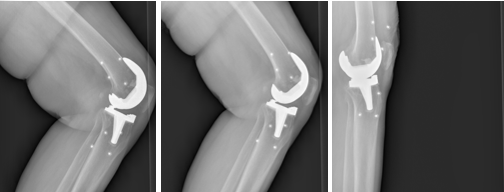}
        \caption{Knee extension motion frames from 2D scanning}
        \label{fig:2D visualization results of extension motion frames}
    \end{subfigure}
    \begin{subfigure}{1\linewidth}
        \centering
        \includegraphics[width=\textwidth]{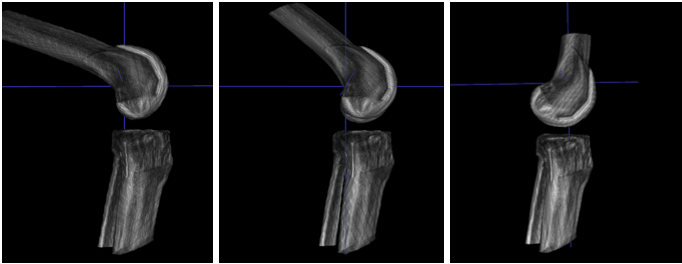}
        \caption{Knee extension motion frames from 4D virtual imaging}
        \label{fig:3D visualization results of extension motion frames}
    \end{subfigure}
    \caption{Visualization of knee extension motion frames}
    \label{fig:distance_distributions}
\end{figure}

Figure~\ref{fig:TP Condyle contact points sub1} illustrates the distribution of tibial plateau (TP)–condyle contact distances under different extension angles measured Jennifer's method \cite{Scarvell2004COMPARISONOK}, where the horizontal and vertical axes represent the medial and lateral distances, respectively. Most points lie below the diagonal (equal medial–lateral distances), indicating asymmetry in TP–condyle contact that supports the presence of leg axis malalignment.  

As supplementary evidence, Figure~\ref{fig:media-lateral difference sub1} shows the medial–lateral distance difference across frames. Here, the horizontal axis corresponds to frame index and the vertical axis to the medial–lateral difference. The maximum observed deviation exceeds 2 mm, a clinically recognized threshold for diagnosing knee malalignment. The medial-lateral distance can be further visualized by the Figure \ref{fig:Extension angle VS medial/lateral distance sub1}, showing the medial/lateral distance changes by extension angles.

Knee joint instability can also be visualized by Feng's method \cite{4DCT_KinematicAnalysis_FlexionExtension}. As shown in Figure~\ref{fig:trajectory_analysis}, trajectories are defined by connecting the medial and lateral condyle projection points on the tibial plateau across frames. Color gradients from light to dark denote progression from the initial to the final frame. Notably, the connecting lines shorten as the joint approaches slight hyperextension (Figure~\ref{fig:hyperextension_frames}), a finding corroborated by both the 2D view (Figure~\ref{fig:2D visualization results of extension motion frames}) and the 3D CBCT reconstruction (Figure~\ref{fig:3D visualization results of extension motion frames}).  

\newpage
\section{Discussion}

This study presents a comprehensive 4D virtual imaging framework that integrates hardware innovation with advanced algorithmic design, offering a feasible approach for dynamic joint evaluation and the early diagnosis of post-arthroplasty instability.  

\subsection{\textbf{Technical advantages}}
The gantry-free, dual–robotic-arm CBCT system enables weight-bearing imaging in physiological postures such as standing or controlled motion \cite{Brinch2022,Li2025}. The reverse-spiral scanning trajectory minimizes motion artifacts from robotic arm vibration, ensuring high-fidelity 3D reconstructions that provide reliable input for subsequent 2D–3D registration.  


On the software side, the 4D virtual imaging framework integrates a three-stage registration pipeline—\textbf{global optimization}, the \textbf{kinematic priority module (KPM)}, and \textbf{hybrid Powell–Nelder-Mead refinement}—to balance accuracy and computational efficiency. Coupled with hardware innovation, this constraint-driven approach defines a new paradigm for dynamic 3D orthopedic imaging, enabling precise characterization of joint mechanics under real-world conditions.  



\subsection{\textbf{Clinical Implications}}  

4D kinematic analysis enables precise characterization of three-dimensional displacement and stress transmission during flexion–extension and weight-bearing. Leveraging virtual 4D imaging and derived kinematic parameters, we propose a clinical framework encompassing three domains: \textit{cartilage integrity}, \textit{ligament balance}, and \textit{joint alignment}.  

\subsubsection{\textbf{Cartilage Integrity and Bone Contact}}  
Static joint spacing (e.g., standing or fixed flexion) provides baseline references, while continuous tracking during motion reveals dynamic femur–tibia interactions. Deviations from physiological patterns highlight early cartilage wear, instability, or implant loosening.  

\subsubsection{\textbf{Ligament Balance and Malalignment}}  
Medial–lateral joint space symmetry reflects ligament function. Although the lateral compartment is physiologically narrower, intercompartmental differences greater than 2 mm indicate dysfunction (e.g., medial collateral laxity or lateral collateral tension), predisposing to cartilage and meniscal injury. Malalignment, such as varus or valgus, further alters load distribution and accelerates degeneration.  

\subsubsection{\textbf{Multi-dimensional Kinematic Metrics}}  
\begin{figure}
    \centering
    \includegraphics[width=1\linewidth]{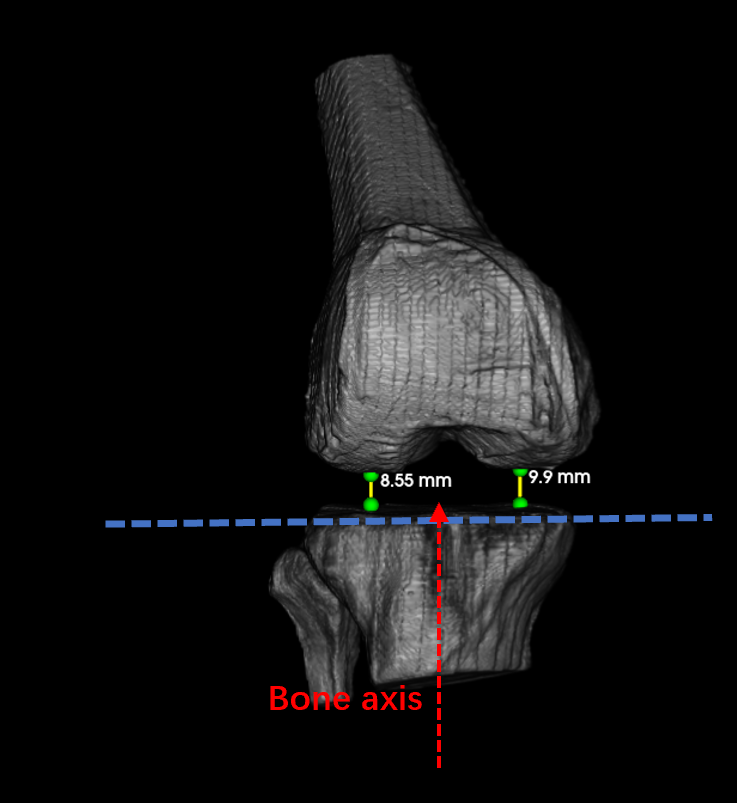}
    \caption{Illustration of femorotibial joint distances measurement}
    \label{fig:tibial_plateau}
\end{figure}

Femorotibial stability was quantified from condylar trajectories projected onto a standardized tibial plateau plane (Figure~\ref{fig:tibial_plateau}). The femorotibial distance was defined as the perpendicular distance between this plane—aligned with the polyethylene insert in post-TKA cases or 9 mm below the distal tibia in pre-TKA cases—and the lowest condylar points. Variations in medial–lateral trajectory morphology across motion frames provide objective markers of malalignment.  



Clinically, this framework bridges static anatomical imaging with dynamic functional assessment, enabling accurate 3D kinematic analysis from a single-plane fluoroscopic monitor while minimizing hardware complexity and radiation exposure. It allows detection of subtle instabilities (e.g., ligament insufficiency), abnormal contact points (e.g., impingement), and supports functional surgical planning under weight-bearing conditions. In knee applications, it quantifies medial–lateral instability and trajectory irregularities across frontal and axial planes. Beyond post-operative TKA stability monitoring, it also detects early osteoarthritic spacing changes. Automated segmentation enhances reproducibility and scalability, positioning this method for multi-center studies and large-scale clinical trials.

\subsection{\textbf{Limitations and Future Directions}}
Several limitations warrant consideration. First, the current clinical dataset is relatively small, limiting generalizability, particularly in complex pathologies. Expanding sample size to include broader demographics and disease severities will be essential. Second, the framework currently emphasizes osseous structures without fully accounting for soft tissue interactions such as ligamentous laxity or cartilage degeneration, which may affect motion accuracy. Incorporating multimodal imaging (e.g., MRI for soft tissue characterization) or physics-based biomechanical modeling could address this limitation. Third, although the KPM improves efficiency, the three-stage registration remains computationally demanding for real-time use. Future work may explore GPU acceleration, parallelization, or lightweight neural network surrogates to enhance processing speed.  

Looking ahead, this system could be adapted to other joints such as the hip or spine and integrated with surgical navigation for intraoperative guidance. Combining 4D kinematic datasets with predictive machine learning models may further enable pre-surgical simulation of implant performance, advancing the development of personalized orthopedic care.  

\section{Conclusion}

This study introduces a 4D virtual imaging framework that integrates upright weight-bearing CBCT with a three-stage hybrid registration strategy, achieving sub-voxel accuracy in six-degree-of-freedom alignment. The gantry-free, dual–robotic-arm system captures high-resolution 3D volumes under physiologically loaded postures, enabling precise visualization of joint kinematics. Clinical validation demonstrates its potential for improving diagnosis and treatment planning in knee osteoarthritis, post-traumatic deformities, and post-arthroplasty instability. By combining biomechanical modeling, deep learning segmentation, and temporospatially constrained registration, the framework establishes a robust pipeline for dynamic imaging in surgical workflows. Although limited by a small clinical sample and incomplete soft tissue modeling, this work lays the groundwork for broader applications and real-time integration in musculoskeletal care.

\section*{Acknowledgment}
This study was funded by the National Science Foundation of China (grant number 82472538), the Beijing Natural Science foundation (funding number L254050), and the Beijing Municipal Administration of Hospitals Program (grant number BJRITO-RDP-2023).

\bibliographystyle{ieeetr}
\bibliography{reference}

\end{document}